\documentclass{article}

\usepackage[final,nonatbib]{neurips_2019}

\usepackage[utf8]{inputenc} 
\usepackage[T1]{fontenc}    
\usepackage{hyperref}       
\usepackage{url}            
\usepackage{booktabs}       
\usepackage{amsfonts}       
\usepackage{nicefrac}       
\usepackage{microtype}      
\usepackage[table]{xcolor}
\usepackage{multirow}
\usepackage{amsmath}
\usepackage{algorithm}
\usepackage{graphicx}  
\usepackage[noend]{algpseudocode}
\usepackage{array}

\title{AutoML using Metadata Language Embeddings}

\author{%
  Iddo Drori \\
  Columbia University and Cornell University \\
  \texttt{idrori@cs.columbia.edu,idrori@cornell.edu} \\
  \And
  Lu Liu \\
  Columbia University \\
  \texttt{ll3252@columbia.edu} \\
  \And
  Yi Nian \\
  Columbia University \\
  \texttt{yn2336@columbia.edu } \\
  \And
  Sharath C. Koorathota \\
  Columbia University \\
  \texttt{sk4172@cumc.columbia.edu } \\
  \And
  Jie S. Li \\
  Columbia University \\
  \texttt{jsl2267@columbia.edu } \\
  \And
  Antonio Khalil Moretti \\
  Columbia University \\
  \texttt{amoretti@cs.columbia.edu } \\
  \And
  Juliana Freire \\
  New York University\\
  \texttt{juliana.freire@nyu.edu } \\
  \And
  Madeleine Udell \\
  Cornell University \\
  \texttt{udell@cornell.edu } \\
  }

\begin{document}

\maketitle

\begin{abstract}
As a human choosing a supervised learning algorithm, it is natural to begin by reading a text description of the dataset and documentation for the algorithms you might use. We demonstrate that the same idea improves the performance of automated machine learning methods. We use language embeddings from modern NLP to improve state-of-the-art AutoML systems by augmenting their recommendations with vector embeddings of datasets and of algorithms. We use these embeddings in a neural architecture to learn the distance between best-performing pipelines. The resulting (meta-)AutoML framework improves on the performance of existing AutoML frameworks. Our zero-shot AutoML system using dataset metadata embeddings provides good solutions instantaneously, running in under one second of computation. Performance is competitive with AutoML systems OBOE, AutoSklearn, AlphaD3M, and TPOT when each framework is allocated a minute of computation. We make our data, models, and code publicly available.
\end{abstract}

\section{Introduction}

A data scientist facing a challenging new supervised learning task
does not generally invent a new algorithm. Instead, they consider what they know about the dataset and which
algorithms have worked well for similar datasets in past experience. Automated machine learning (AutoML) seeks to automate these tasks to enable widespread use of machine learning by non-experts. A major challenge is to develop \emph{fast, efficient} algorithms to accelerate applications of machine learning \cite{kokiopoulou2019fast}.
This work develops automated solutions that exploit human expertise to learn \emph{which datasets are similar} and 
\emph{what algorithms perform best}. We use modern natural language processing (NLP) tools to teach AutoML systems how to read text descriptions of datasets, and develop a structured representation of the solutions to Kaggle challenges to allow our system to run winning solutions on new datasets.

A simple idea is to use machine learning pipelines that performed well (for the same task) on similar datasets.
What constitutes a similar dataset? The success of an AutoML system often hinges on this question,
and different frameworks have different answers: for example, AutoSklearn \cite{feurer2015efficient} computes a set of \emph{metafeatures} for each dataset, while OBOE \cite{yang2019oboe} uses the performance of a few fast, informative models to compute latent features. More generally, for any supervised learning task, one can view the list of recommended algorithms generated by any AutoML system as a vector describing that task. Oddly, no previous work uses the information that a human would check first: a summary description of the dataset, written in free text. These dataset features induce a metric structure on the space of datasets. Under an ideal metric, a model that performs well on one dataset would also perform well on nearby datasets. The methods we develop in this work show how to learn such a metric using the recommendations of \emph{any} AutoML framework together with the dataset description.

This work makes several contributions to the literature
by marrying techniques from NLP with AutoML.
First, we develop NLP text embeddings for datasets by reading the dataset metadata,
including dataset title, description, and keywords.
We show that using these embeddings improves the performance of
state of the art AutoML frameworks such as 
OBOE \cite{yang2019oboe}, AutoSklearn \cite{feurer2015efficient}, AlphaD3M \cite{drori2019alphad3m}, and TPOT \cite{olson2019tpot}. Second, we develop NLP text embeddings for machine learning pipelines by reading the algorithm documentation. We show how to use these embeddings to develop a new training objective for AutoML:
for any two training datasets, we compute the distance between the embeddings of the algorithm that performs best on each dataset. We learn a metric on dataset embeddings to match the distance between the embeddings of the corresponding best algorithms. We can use this metric for \emph{zero-shot AutoML}: given a new dataset,
we compute its embedding (given a text description of the dataset), use it to find the closest training dataset, and output the best algorithm known for that training dataset. Using the additional information present in the dataset metadata embeddings and pipeline embeddings improves performance of existing AutoML systems. The third major contribution of this work is a new metadata dataset for AutoML that we call AutoKaggle. AutoKaggle consists of a collection of Kaggle competitions, tasks, winning pipelines, and an execution engine.

AutoML is an emerging field of machine learning with the potential to transform the practice of data science by automatically choosing a model to best fit the data. The reader interested in a comprehensive review of the field can consult one of the three surveys published in the last twelve months \cite{yao2018survey,he2019automl,zoller2019survey}. A new benchmark \cite{amlb2019} provides a quantitative comparison of many top algorithms. 

Language has a common unstructured representation of words, sentences, paragraphs, and trees of paragraphs which form stories. The most significant recent advances in NLP learn language models and embeddings from very large corpuses of text \cite{devlin2018bert,radford2019language}. An unsupervised corpus of text is transformed into a supervised dataset by defining content-target pairs along the entire text: for example, target words that appear in each sentence, or target sentences which appear in each paragraph. A language model is first trained to learn a low dimensional embedding of words or sentences followed by a map from low dimensional content to target \cite{mikolov2013efficient}. This embedding can be used to embed on a new, unseen and small, dataset in the same low-dimensional space. This work is the first to propose using such embeddings for automatic machine learning. 
Specifically, we use an embedding for the datasets, an embedding for the pipelines, and the non-linear interactions between these embedding using a neural network.

One major factor in the performance of an AutoML system is the base set of algorithms it can use to compose more complex pipelines. For a fair comparison, in our numerical experiments we compare our proposed methods only to other AutoML systems that build pipelines out of Scikit-learn \cite{scikit-learn} primitives. Now, humans who compete in Kaggle competitions do not restrict themselves in the same way. Hence as part of AutoKaggle, we have developed and released translations of every winning Kaggle entry that comply with the AutoKaggle pipeline format. Each of these can be interpreted by the AutoKaggle execution engine. The resulting Scikit-learn translation of the pipeline is sometimes better, but generally a tad worse, than the original human-engineered pipeline.

\section{Methods}

\begin{table*}
\small
\begin{tabular}{ll}
 \textbf{Notation} & \textbf{Description} \\
 \hline
 $\mathcal{D}$ & Dataset\\
 $\mathcal{M_{D}}$ & Metadata of dataset $\mathcal{D}$ \\
 \hline 
 $\mathcal{T}$ & Machine learning task (classification, regression)\\
 \hline
 $\mathcal{P}$ & Solution pipeline \\
 $\mathcal{O, S, A, G, H}$ & OBOE, AutoSklearn, AlphaD3M, TPOT, and human algorithm \\
 $\mathcal{P_{B}(D,T)}$ for $\mathcal{B} \in \{\mathcal{O, S, A, G, H}\}$ & Solution pipeline on dataset $\mathcal{D}$ for task $\mathcal{T}$ \\
$\mathcal{V(P_{B}, D, T)}$ & Evaluating performance of pipeline $\mathcal{P_{B}}$ on $\mathcal{D}$ and $\mathcal{T}$\\
 \hline
 $E$ & Pre-trained language embedding \\
 \hline
 $E(\mathcal{M_{D}})$ & Language embedding of dataset metadata\\
 $d(E(\mathcal{M}_{\mathcal{D}_{i}}), E(\mathcal{M}_{\mathcal{D}_{j}}))$ & Distance between dataset metadata embeddings\\
 $\mathcal{D_{\star}} = \underset{\mathcal{D}_{i}}{\text{argmin}} \|E(\mathcal{M}_{D}), E(\mathcal{M}_{D_{i}})\|$ & Nearest neighbor of $\mathcal{D}$ under distance of embeddings \\
 $\mathcal{P_{\star} = P(D_{\star}, T)}$ & Pipeline of most similar embedding \\
 $\mathcal{V(P_{\star}, D, T)}$ & Direct pipeline transfer using dataset metadata embedding \\
 \hline
 $E(\mathcal{P_{B}(D,T)})$ & Language embedding of solution pipeline \\
 $\mathcal{X(D,T)}$ & Representation of embeddings for dataset $\mathcal{D}$ and task $\mathcal{T}$\\
 \hline
 Interaction between embeddings &\\
 $\mathcal{I}(\mathcal{D}_{i}, \mathcal{D}_{j}) = (\mathcal{X}(\mathcal{D}_{i},T)$, $\mathcal{X}(\mathcal{D}_{j},\mathcal{T}))$ & Neural network input: pair of representations $\mathcal{X(D,T)}$\\
 $\mathcal{O}(\mathcal{D}_{i}, \mathcal{D}_{j}) = d(E(\mathcal{P_{H}}(\mathcal{D}_{i},\mathcal{T})), E(\mathcal{P_{H}}(\mathcal{D}_{j},\mathcal{T}))$ & Network output: distance between human pipeline embeddings \\
 \hline
\end{tabular}
    \caption{Embedding AutoML notation and their descriptions.}
    \label{table:notation}
\end{table*}

This work uses NLP embeddings to find machine learning pipelines that perform well for a given dataset and task. We separately embed dataset metadata and machine learning pipelines and pass the embeddings through a neural network. This work designs appropriate embedding methods for both dataset metadata and machine learning pipelines and demonstrates that the resulting recommender system works well. Table \ref{table:notation} provides the mathematical notation that defines these methods.

We rely on NLP tools to produce embeddings of dataset metadata $\mathcal M_{\mathcal D}$
and of pipelines. Concretely, in our experiments, we embed dataset metadata by applying the USE3 embedding \cite{cer2018universal} to the dataset description (including title, subtitle, description, and keywords) to form $E(\mathcal M_{\mathcal D})$, and embed a pipeline $\mathcal P$ by applying the embedding to the function call and the header for each estimator used in the pipeline to form $E(\mathcal P)$.

To facilitate evaluation of arbitrary pipelines, we have developed an \emph{execution engine} in Python that can represent pipelines composed of machine learning primitives from the Scikit-learn library. The execution engine takes as input a description of a pipeline $\mathcal P$ (consisting of machine learning primitives and their parameters, structured in a chain describing the order of execution) and computes the value $\mathcal{V}(\mathcal P, \mathcal D, \mathcal T)$ of pipeline $\mathcal{P}$ on task $\mathcal{T}$ for dataset $\mathcal{D}$ by running the primitives sequentially on a given dataset $\mathcal D$ and task $\mathcal T$.

\begin{figure*}[t!]
    \centering
    \includegraphics[width=0.95\textwidth]{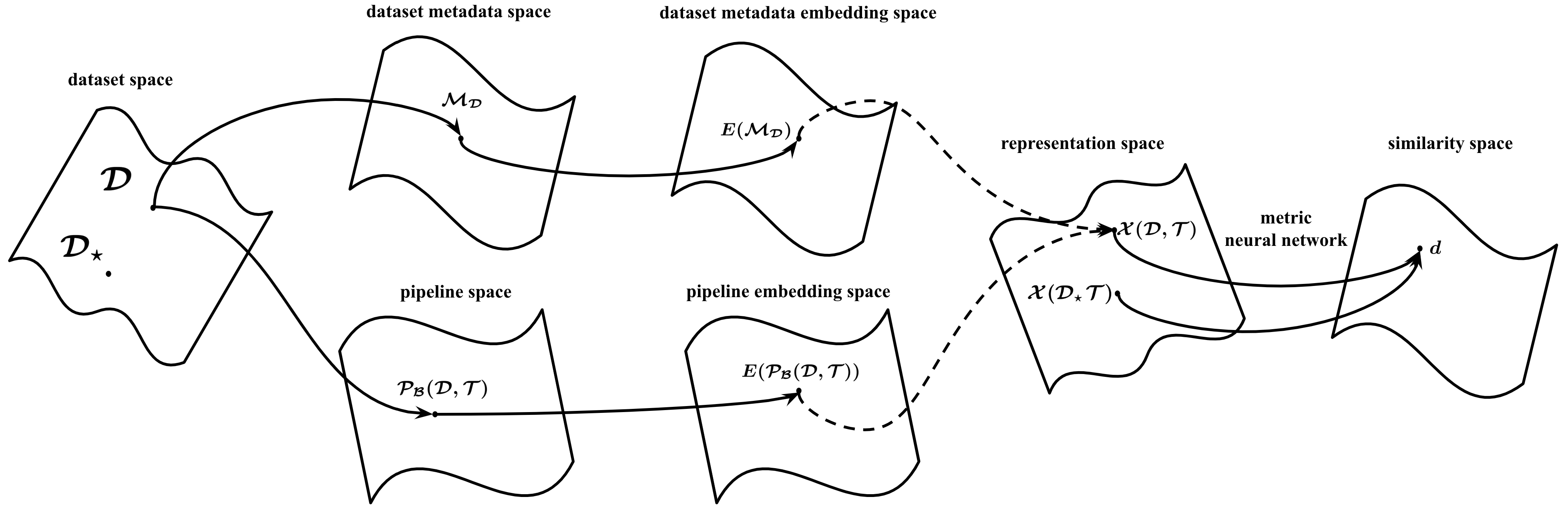}
    \caption{AutoML embeddings of dataset metadata and pipelines. The dashed arrows denote that the representation may consist of any number of the embeddings. }
    \label{fig:automl-embeddings}
\end{figure*}

The execution engine allows us to run any pipeline for a given task on any dataset. Our hypothesis is that datasets whose metadata embeddings are similar will share successful pipelines. To test this hypothesis, we develop an AutoML approach that we call \emph{direct pipeline transfer}. For a given dataset $\mathcal{D}$, we find the most similar dataset with respect to the metadata embedding:
\begin{equation}
\mathcal{D}_{\star} = \underset{\mathcal D'}{\text{argmin}} \|E(\mathcal M_{\mathcal D}) - E(\mathcal M_{\mathcal D'})\|,
\end{equation}
and evaluate the corresponding pipeline $\mathcal{P_{\star}}$ on the original dataset $\mathcal{D}$ to compute $\mathcal{V(P_{\star}, D, T)}$. Dataset metadata is useful alone, but can be even more powerful in combination with other information. We develop embeddings $\mathcal{X(D,T)}$ that are formed by concatenating metadata embeddings $E(\mathcal M_{\mathcal D})$ with pipeline embeddings $E(\mathcal{P(D,T)})$ for pipelines produced by any AutoML system, as shown in Figure \ref{fig:automl-embeddings}. We refer to these dataset embeddings as AutoML embeddings. We extend the direct pipeline transfer methodology to use these more complex dataset representations. For a given dataset $\mathcal{D}$ with representation $\mathcal{X(D,T)}$, we find the most similar dataset with respect to the metadata embedding: 
\begin{equation}
\mathcal{D}_{\star} = \underset{\mathcal D'}{\text{argmin}} \|\mathcal{X(D,T)} - \mathcal{X(D',T)}\|,
\end{equation}
and evaluate the corresponding pipeline $\mathcal{P}_{\star}$ on the original dataset $\mathcal{D}$ to compute $\mathcal{V(P_{\star}, D, T)}$. A disadvantage of this method is that we must specify both the similarity metric and the importance of each component of the representation. 
Instead, we can learn the similarity metric and interaction between representations $\mathcal{X(D,T)}$ automatically using a neural network, to learn the similarity metric and output the distance between representations. The neural network takes as input pairs of datasets $\mathcal D_i$ and $\mathcal D_j$ and is trained to output 
the performance of the human-selected model for dataset, $\mathcal{P_{H}}(\mathcal{D}_{i},\mathcal{T})$,
evaluated on dataset $\mathcal D_j$:
$\mathcal{V}(\mathcal{P_{H}}(\mathcal{D}_{i},\mathcal{T}), \mathcal{D}_j, \mathcal{T})$.
At prediction time, given a new dataset $\mathcal{D}$, we first compute the representation $\mathcal{X}(\mathcal{D},\mathcal{T})$, and then compute the distance of the new dataset to all other datasets using a neural network. We choose the pipeline $\mathcal P$ corresponding to the dataset with the smallest distance, and evaluate its performance $\mathcal{V}(\mathcal{P}, \mathcal{D}, \mathcal{T})$ on the new dataset $\mathcal{D}$.

An important methodology in data science is the common task framework \cite{blei2017datascience,donoho2017fifty} in which a common dataset and task is given to multiple participants and evaluated using the same performance metrics. In this work we curate AutoKaggle, a meta-dataset which contains meta-data about datasets $\mathcal{D}$, tasks $\mathcal{T}$, and solution pipelines $\mathcal{P}$. This dataset is important for: analyzing which solution components are used for which datasets and tasks, understanding which tasks and sub-tasks are given for which datasets, recommending which high performance solutions be used for new unseen datasets and tasks, and identifying usage trends of machine learning libraries and primitives. Our new meta-dataset contains structured information about a wide variety of machine learning tasks, together with meta-data about the data, task, and solution pipelines. Solution source code is parsed into structured machine learning pipelines including: pre-processing operations, feature extractors, feature selectors, estimators, and post-processing operations. The dataset can be viewed as a sparse high dimensional tensor. Rows correspond to (dataset, problem) pairs; other dimensions correspond to possible values for preprocessors, feature extractors, feature selectors, estimators, and post-processors. The entries of the tensor are the performance of the corresponding pipeline for the task on the dataset.

\section{Results}
Table \ref{tab:pipeline-evaluations} shows our results for a representative set of tabular datasets on classification tasks. For each dataset (row), Table \ref{tab:pipeline-evaluations} reports mean evaluation accuracy of different pipelines (columns) running on the same well-defined task. Specifically, prediction accuracy of OBOE $\mathcal{V(P_{O}(D,T))}$, AutoSklearn $\mathcal{V(P_{S}(D,T))}$, AlphaD3M $\mathcal{V(P_{A}(D,T))}$, and TPOT $\mathcal{V(P_{G}(D,T))}$, evaluation of human generated pipeline $\mathcal{V(P_H(D,T))}$, and the predicted pipeline accuracy of the best dataset metadata embedding $E(\mathcal{M_{D}})$ and single pipeline embedding $E(\mathcal{P_{B})}$. All AutoML systems were given one minute of computation time for a fair comparison; whereas our zero-shot AutoML using dataset metadata embedding $E(\mathcal{M_{D}})$ runs under one second of computation, and our pipeline embedding $E(\mathcal{P_{B})}$ runs within the same one minute of computation while improving performance. 

\begin{table*}[h]
\centering
\small
\begin{tabular}{lccccccc}
Dataset & OBOE & AutoSklearn & AlphaD3M & TPOT & Human & \textbf{Ours DE} & \textbf{Ours PE}\\
\hline
Seattle & \textbf{1.00} & 0.66 & \textbf{1.00} & \textbf{1.00} & 0.92 &  \textbf{1.00} & \textbf{1.00}\\
Insurance & 0.47 & 0.50 & 0.35 & 0.51 & 0.48 & 0.47 &  \textbf{0.52}\\
Forest & 0.73 & 0.83 & 0.83 & 0.84 & 0.84 & \textbf{0.85} & \textbf{0.85}\\
Credit & 0.93 & \textbf{0.94} & 0.93 & \textbf{0.94} & \textbf{0.94} & \textbf{0.94} & \textbf{0.94}\\
Titanic & 0.70 & 0.80 & 0.70 & 0.77 & 0.86 & \textbf{0.87} & \textbf{0.87}\\
HR & 0.84 & 0.83 & 0.87 & \textbf{0.90} & 0.86 & 0.86 & \textbf{0.90}\\
Kobe & 0.61 & 0.60 & \textbf{0.64} & 0.62 & 0.62 & 0.61 & 0.62\\
Patients & 0.64 & 0.71 & \textbf{0.72} & 0.68 & 0.68 & 0.68 & 0.69\\
\hline
\end{tabular}
\caption{Machine learning pipeline evaluations for AutoML systems and human pipelines. All AutoML pipelines $\mathcal{P_{O}, P_{S}, P_{A}, P_{G}}$ are computed given 1 \emph{minute} of computation. In comparison, ours DE refers to $E(\mathcal{M_{D}})$ using only the dataset metadata embedding in under 1 \emph{second} of computation for zero-shot AutoML. Ours PE refers to $E(\mathcal{P_{B}})$ using the best single pipeline embedding $\mathcal{B} \in \{\mathcal{O, S, A, G}\}$ in 1 minute of computation.}
\label{tab:pipeline-evaluations}
\end{table*}

To implement the metric neural network that learns to predict the distance $d(E(\mathcal{P_{H}}(\mathcal{D}_{i},\mathcal{T})), E(\mathcal{P_{H}}(\mathcal{D}_{j},\mathcal{T})))$ between the predicted pipeline embedding of pairs of datasets $\mathcal{D}_{i}$ and $\mathcal{D}_{j}$, 
we construct a fully connected neural network with four layers, batch size of 16, and 1200 training epochs. We use the Adam optimizer with 0.001 learning rate. The input to the neural network is the representation $\mathcal{X(D,T)}$ of the test dataset and the representation $\mathcal{X}(\mathcal{D}_{i},\mathcal{T})$ of every other dataset. We train the neural network for every test dataset and get our evaluation accuracy by running the obtained pipeline on the test dataset using our execution engine.

\section{Conclusions}

We have introduced a neural architecture to embed textual descriptions of dataset metadata and machine learning pipelines for AutoML. We use a new dataset AutoKaggle consisting of structured representations of winning solutions of Kaggle competitions and an execution engine to run multiple ML pipelines. We make our data, models, and code publicly available \cite{autommlembeddings2019code}. In future work we would like to apply our method to additional AutoML systems such as Auto-WEKA and H2O AutoML. We would also like to compare the performance of embedding AutoML using different large language embeddings such as BERT \cite{devlin2018bert} and GPT-2 \cite{radford2019language}.

\bibliographystyle{plain}
\bibliography{bibliography}

\end{document}